\pdfoutput = 1
\documentclass[10pt]{article}
\usepackage{spconf,amsmath,graphicx}

\begin{document}
\title{Speech Recognition with no speech or with noisy speech} 
\name{Gautam Krishna$^{\star}$ \qquad Co Tran$^{\star \ddagger}$ \qquad Jianguo Yu$^{\dagger \ddagger}$ \qquad Ahmed H Tewfik$^{\star}$ \thanks{$\ddagger$ Equal author contribution}}

\address{$^{\star}$Brain Machine Interface Lab, The University of Texas at Austin \\
$^{\dagger}$ University of Aizu\\}

\author{
\IEEEauthorblockN{Gautam Krishna}
\IEEEauthorblockA{Department of Electrical and\\Computer Engineering\\
 University of Texas at Austin}
\and
\IEEEauthorblockN{Co Tran*}
\IEEEauthorblockA{Department of Mathematics \\College of Natural Sciences\\
 University of Texas at Austin}
\and
\IEEEauthorblockN{Jianguo Yu*}
\IEEEauthorblockA{Human Interface Lab\\University of Aizu\\
Fukushima,Japan}
\and
\IEEEauthorblockN{Ahmed H Tewfik}
\IEEEauthorblockA{Department of Electrical and\\Computer Engineering\\
 University of Texas at Austin}
 \thanks{denotes equal author contribution}
}%
\maketitle
\begin{abstract}
\textbf{The performance of automatic speech recognition systems(ASR) degrades in the presence of noisy speech. This paper demonstrates that using electroencephalography (EEG) can help automatic speech recognition systems overcome performance loss in the presence of noise. The paper also shows that distillation training of automatic speech recognition systems using EEG features will increase their performance. Finally, we demonstrate the ability to recognize words from EEG with no speech signal on a limited English vocabulary with high accuracy.
}
\end{abstract}

\section{{\bf \uppercase{Introduction}}}
\label{intro}
Traditional state of art Automatic Speech Recognition (ASR) systems mainly uses acoustic features for doing speech recognition.
In \cite{kirchhoff2002combining} authors show how combining acoustic and articulatory features can help in designing robust speech recognition systems. Recently, researchers have used Functional near-infrared spectroscopy (fNIRS) signals for doing speech recognition with 74.7 \% accuracy \cite{liu2018speech}.
In \cite{ramsey2017decoding,martin2016word} authors provide interesting results on how Electrocorticography (ECoG) signals, which is an invasive approach can help in speech recognition. 
\\ 
Electroencephalography (EEG) on other hand is a non invasive approach. It is a measure of electrical activity of the human brain.
In \cite{yoshimura2016decoding} authors demonstrate decoding vowel articulation using EEG cortical currents. In our work we used only surface EEG potentials, which are directly obtained from the EEG sensors.\\  
In \cite{kumar2018envisioned} authors used EEG signals to perform envisioned speech recognition using random forest algorithm and they reported an average accuracy of 85.2 \%. In our work we used a deep learning model and achieved a highest test accuracy of 99.38 \%. In \cite{sun2016neural} authors propose neural network based model which predicts phonemes from EEG but in our work the model directly predicts words with higher accuracy and we also study the effect of noisy speech.
References \cite{sriram2018robust,mcloughlin2015robust,gemmeke2011exemplar,tan2018adaptive} describes various techniques to perform speech recognition with noisy speech but as far as we know our work is first demonstration of EEG's ability to overcome performance loss in presence of background noise and our approach demonstrates a significant high recognition accuracy of 99.38 \% for recognition of limited words in presence of background noise. 

In \cite{hinton2014distilling} Hinton proposed the concept of distilling the knowledge in neural networks, where a simple model learns a complex task by imitating the solution of a more complicated and flexible model.
In \cite{yu2016articulatory} authors demonstrate the integration of acoustic and articulatory features using Generalized Distillation.\\ 
Motivated by the results presented in \cite{liu2018speech,sun2016neural,schultz2017biosignal,yoshimura2016decoding,alsaleh2016brain,kumar2018envisioned,rosinova2017voice,kim2014eeg},  primary goal of our research was to create a state of art ASR system and train it with EEG features, acoustic features, concatenation of acoustic - EEG features and investigate its performance in absence and presence of background noise. We tested our model for recognition of the five English vowels and four English words 'yes', 'no' , 'left', 'right'.\\ 
Inspired from \cite{hinton2014distilling} and \cite{yu2016articulatory} we further tried implementing Generalized Distillation in the speech recognition task in order to integrate EEG information into speech recognition system and observed that distillation training with EEG improves the recognition accuracy of our ASR model. \\

Major contributions of our paper are as follows: we identified a set of EEG features which are better representation of speech, we proposed a deep learning model that is capable of learning EEG features and perform speech recognition with no speech as input, we demonstrate the ability of EEG features to make up for ASR performance lost due to background noise and we show that performance of ASR system can be improved by distillation training with EEG features.
\section{{\bf \uppercase{Automatic Speech Recognition System Model}}}
For this work we created an ASR model using gated recurrent unit (GRU) networks \cite{chung2014empirical}. The model was created using Google TensorFlow deep learning library. GRU has an architecture similar to long short term memory (LSTM) but has less parameter's compared to LSTMs. Hence GRU's are ideal for recurrent neural network applications where less amount of training data is available.
\begin{figure}[h]
\includegraphics[height=3cm, width=0.5
\textwidth,trim={0.1cm 0.1cm 0.1cm 0.1cm},clip]{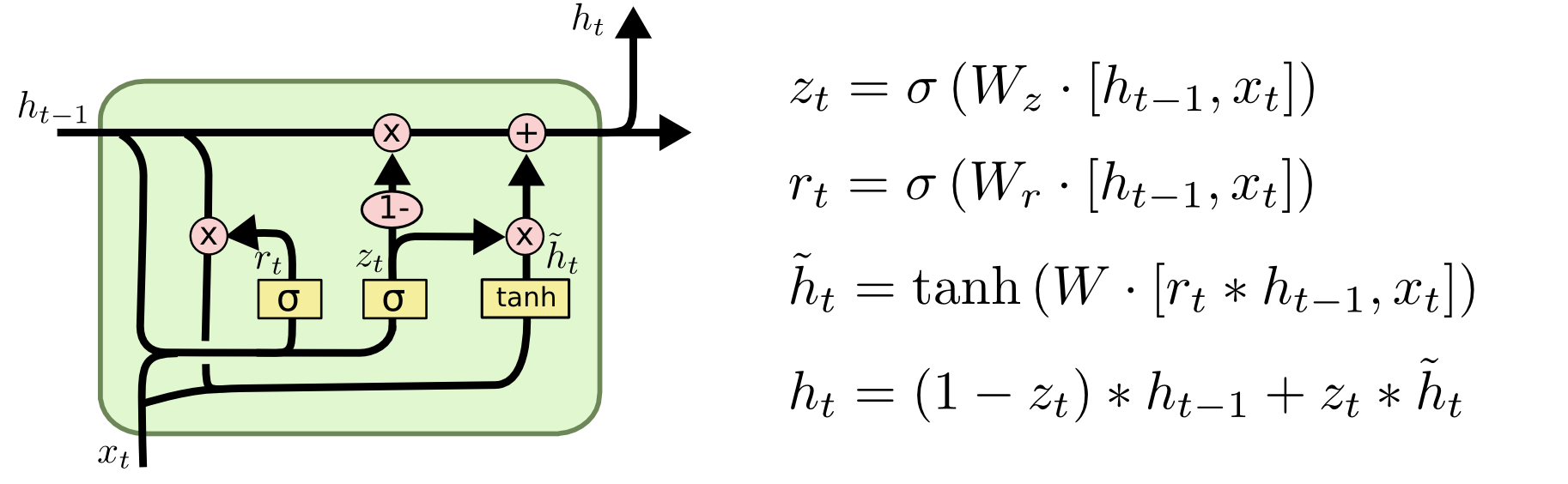}
\caption{Gated Recurrent Unit Cell}
\label{1vsall}
\end{figure}
A GRU cell architecture is shown in Figure 1, $x_t$ is the input vector, $h_t$ is the output vector, $z_t$ is update gate vector and $r_t$ is reset gate vector. Sigmoid and hyperbolic tangent activation functions are used in the GRU cell.\\ 
Given the sequence of input vectors $X = [x_0, x_1, ..., x_n]$ and $x_i \in R^{d}$ is the input vector at time $i$. Our model contains a GRU layer, an average pooling layer and a dense (fully connected) layer followed by output layer, which takes $X$ and returns the probabilities being the all words (vowels) $o \in R^p$, as shown in Figure 2. The pooling layer computes the average value $\frac{1}{n} \sum_{i=0}^n h_i$ of all outputs of GRU layer.


The number of hidden units in GRU layer is 128 and in dense layer is 64 respectively, the output dimension $p$ is 4 or 5 corresponding to the number of classes. The batch size is 1 and the dropout rate for the dense layer is 0.2. We used Adam Optimizer with 0.001 learning rate.
 We used cross entropy as the loss
function.

\begin{figure}[h]
\label{fig:asrmodel}
\includegraphics[height=5cm, width=\linewidth,trim={0.1cm 0.1cm 0.1cm 0.1cm}, clip]{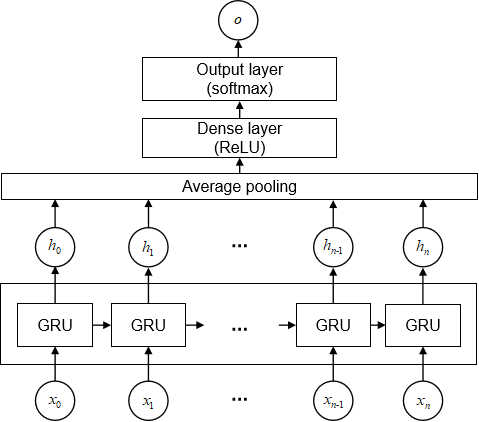}
\caption{Our ASR Model Architecture} 
\label{1vsall}
\end{figure}

\begin{table}
\label{table:training}
\begin{center}
 \begin{tabular}{|p{1.5cm}|p{1cm}|p{1cm}|p{1.5cm}|p{0.6cm}|p{0.5cm}|} 
 \hline
 Words/Vowels & Class & Training set & Validation set & Test set & Total \\ [0.5ex] 
 \hline\hline
  & Ratio & 64 & 16 & 20 & 100 \\ 
 \hline
 Words & yes & 195 & 49 & 61 & 305 \\
 \hline
 Words & no & 259 & 66 & 81 & 406 \\
 \hline
 Words & right & 219 & 56 & 68 & 343 \\
 \hline
 Words & left & 214 & 54 & 67 & 335 \\\hline
 Vowel & a & 170 & 44 & 53 & 267\\\hline
 Vowel & e & 170 & 44 & 53 & 267\\\hline
 Vowel & i & 170 & 44 & 53 & 267\\\hline
 Vowel & o & 170 & 44 & 53 & 267\\\hline
 Vowel & u & 170 & 44 & 53 & 267\\\hline
\end{tabular}
\\
\caption{Training, Validation and Test sets}
\end{center}
\end{table}

As shown in Table 1, 64 percent of the data was used for training set, 16 percent for validation set and remaining 20 percent for testing set.

\begin{figure}[h]
\begin{center}
\includegraphics[height=3cm,width=0.25\textwidth,trim={1cm 1cm 1cm 0.1cm},clip]{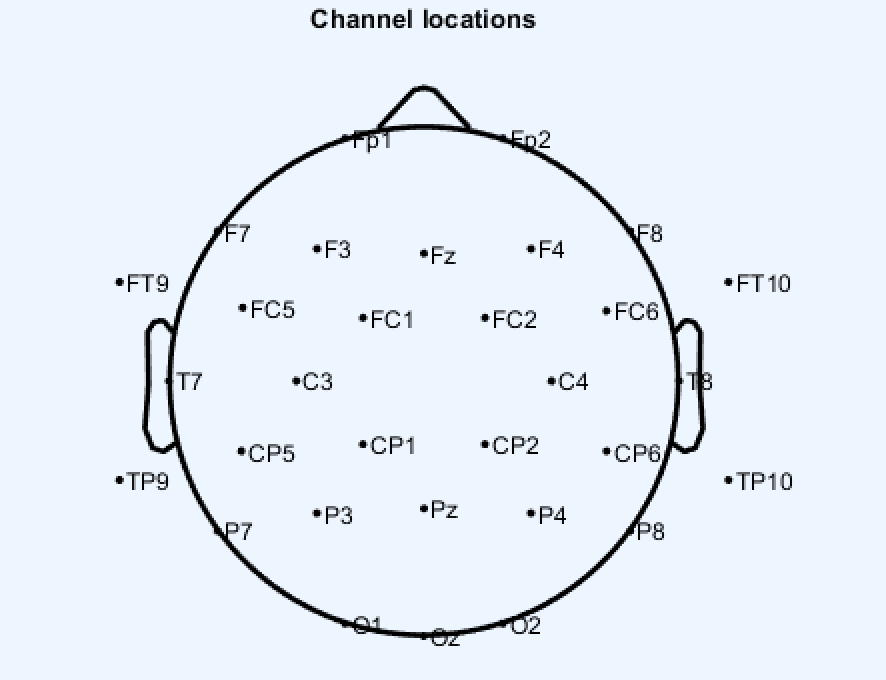}
\caption{EEG channel locations for the cap used in our experiments} 
\label{1vsall}
\end{center}
\end{figure}
We trained and tested GRU based deep learning ASR model using three different feature sets. 1) only acoustic features, 2) concatenation of acoustic and EEG features and 3) only EEG features. 
Number of training epochs was 10000 for all cases except for vowels in presence of noise data set. For that data set the number of epochs was set to 30000 as we didn't observe convergence at 10000 epochs. In Figure \ref{fig:asrmodel} observation vectors $x_0$,$x_1$ up to $x_n$ are treated as input vectors. The index denotes the time step value. There was no fixed time step value for the GRU. We used dynamic recurrent neural network (RNN) cell of tensorflow. Observation vector X can be Mel-frequency cepstral coefficients (MFCC) / acoustic features , EEG or concatenation of EEG and MFCC features depending on how the ASR model was trained. 
\section{\bf \uppercase{GENERALIZED DISTILLATION}}
Distillation training involves following three main steps.\\
1. Train an ASR model with EEG + MFCC. This model is called the teacher model.\\
2. Generate soft targets from this model. After training the teacher model, we used \texttt{estimator.predict} from tensorflow (temperature parameter) to compute the unscaled prediction probabilities for each training example. This unscaled prediction probabilities are called soft targets. \\
3. Train an ASR model with MFCC  + soft targets. This model is called the student model.\\ 
The hyper parameters to tune for the training the student model are temperature and lambda. Temperature is a hyper parameter of neural network used to control the randomness of predictions by scaling the logits (raw predictions) before applying softmax activation function \cite{hinton2014distilling}.
Training loss for the student model is defined as:
\begin{equation}
    \sum(hard\_loss*(1-Lambda)+soft\_loss*(Lambda))
\end{equation}
Where $soft\_loss$ is the cross entropy loss between soft targets,soft logits and $hard\_loss$ is the cross entropy loss between hard targets,logits. 
The parameter lambda behaves like a regularization parameter in the loss function of the student model. It is called the imitation parameter.\\
We tuned the hyper parameters temperature and lambda for the following values:-  Temperature = [1.0,2.0,5.0,10.0] and Lambda =[0,0.2,0.8,1.0] \cite{hinton2014distilling}. We used the ASR model shown in Figure 2 for both our teacher and student model. Instead of the average pooling layer, we used the last time step output of the GRU layer as input to the dense layer for distillation training.
\section{{\bf \uppercase{Design of Experiments for building the database}}}
Four subjects took part in the experiment.All subjects were male undergraduate students in their early twenties. Three were native English speakers and one had an accent. 
The subjects were asked to speak English vowels [a,e,i,o,u] separately for 5 minutes each with a time interval of 2 seconds for each vowel. Their simultaneous speech and EEG signals were recorded. 
The same subjects were then asked to speak the English words [yes,no,left,right]. Their simultaneous speech and EEG signals were recorded. 
The words were spoken for 5 minutes each with a time interval of 5 seconds for each word.\\
We then repeated the same set of  EEG- Speech recording experiments for recognition of English words and vowels in presence of background noise. For generating a background noise of 60 db we used background music played from our lab computer as the source of the noise. 
We used Brain Vision EEG recording hardware. Our EEG cap had 32 wet EEG electrodes including one electrode as \textbf{ground} as shown in Figure 3. We used EEGLab \cite{delorme2004eeglab} to obtain the EEG sensor location mapping. It is based on standard 10-20 EEG sensor placement method for 32 electrodes \cite{sharbrough1991american}.\\ 
In total for this work we used 75 minutes of speech EEG data for vowels with noise, 75 minutes of speech EEG data for vowels without noise, 40 minutes of speech EEG data for words with noise and 40 minutes of speech EEG data for words without noise. The data was recorded from the subjects on different days.\\

\section{\bf \uppercase{EEG and Speech feature extraction details}}
EEG signals were sampled at 1000Hz and a fourth order IIR band pass filter with cut off frequencies 0.1Hz and 70Hz was applied. A notch filter with cut off frequency 60 Hz was used to remove the power line noise.
EEGlab's \cite{delorme2004eeglab} Independent component analysis (ICA) toolbox was used to remove other biological signal artifacts like electrocardiography (ECG), electromyography (EMG), electrooculography (EOG) etc from the EEG signals. 
We extracted five statistical features for EEG, namely root mean square, zero crossing rate,moving window average,kurtosis and power spectral entropy \cite{zhang2008feature}. So in total we extracted 31(channels) X 5 or 155 features for EEG signals.
The EEG features were extracted at a sampling frequency of 100Hz for each EEG channel.\\  
The recorded speech signal was sampled at 16KHz frequency. We extracted MFCC as features for speech signal.
We first extracted MFCC 13 features and then computed first and second order differentials ( delta and delta-delta) thus having total MFCC 39 features.
The MFCC features were also sampled at 100Hz same as the sampling frequency of EEG features to avoid seq2seq problem. 
\section{{\bf \uppercase{EEG Feature Dimension Reduction Algorithm Details}}}
\label{sectionlsh}
After extracting EEG and acoustic features as explained in the previous section, we used non linear methods to do feature dimension reduction in order to obtain set of EEG features which are better representation of acoustic features. 
We reduced the 155 EEG features to a dimension of 39 by applying Kernel Principle Component Analysis (KPCA).We plotted cumulative explained variance versus number of components to identify the right feature dimension. We used KPCA with polynomial kernel of degree 3. 
We further computed delta, delta and delta of those 39 EEG features, thus the final feature dimension of EEG was 117 (39 times 3).
This approach gave best performance for feature dimension reduction for EEG data recorded for words in presence, absence of background noise and for vowels in presence of background noise.
For EEG data recorded for vowels in absence of background noise we used autoencoder for doing feature dimension reduction. Here the EEG feature dimension was first reduced to 6 by autoencoder and delta, delta and delta features were computed thus making the final EEG feature dimension equal to 18 for that data set. 
\section{{\bf \uppercase{Results}}}
The evaluation metric was recognition accuracy. We defined ASR recognition accuracy as the ratio of number of correct predictions given by our model to the total number of predictions given by the model in training,validation and test data sets respectively.\\
\begin{table}
\begin{center}
    \begin{tabular}{|p{2cm}|p{1.75cm}|p{1cm}|p{1.25cm}|p{1cm}|}
         \hline
 Words/Vowels & Background noise & MFCC acc & MFCC-EEG acc & EEG acc \\ 
          \hline\hline
          Vowels & No & 88.75 & \bf{97.50} & 91.25\\
          \hline
          Vowels & Yes & 73.33 & \bf{92.00} & 92.00\\
          \hline
          Words & No & 95.83 & \bf{97.91} & 96.52\\
          \hline
          Words & Yes & 94.53 & \bf{98.39} & 98.39\\
          \hline

    \end{tabular}
    \caption{Validation accuracy for \textbf{ASR EEG fusion} for different datasets}
\end{center}
\end{table}
\\
Table 2 and 3 shows validation accuracy, test accuracy values obtained after convergence for different data sets when trained using acoustic only,concatenation of acoustic and EEG, EEG only features respectively. The test, train and validation accuracy values were comparable indicating that our model didn't over fit. \\
When we trained the model using 31 EEG channels + MFCC, we obtained a test accuracy of 96.36\% on vowels in absence of noise dataset as shown in Table 3. In order to make the system more applicable to real world, we also trained the model with a smaller feature set containing only 4 EEG channels (T7, T8, Fc5, P7) + MFCC and obtained a remarkably close 93\% accuracy on the same dataset.\\
We obtained best results during test time after distillation training for the hyper parameters temperature equal to 2 and lambda equal to 0.2, as shown in Table 4.\\
Table 5 shows test accuracy values obtained after distillation training for different data sets. Student model underwent distillation training but during its test time EEG features are not provided.

\begin{figure}[!ht]
\includegraphics[width=0.42
\textwidth,trim={0.1cm 0.1cm 0.1cm 0.1cm},clip]{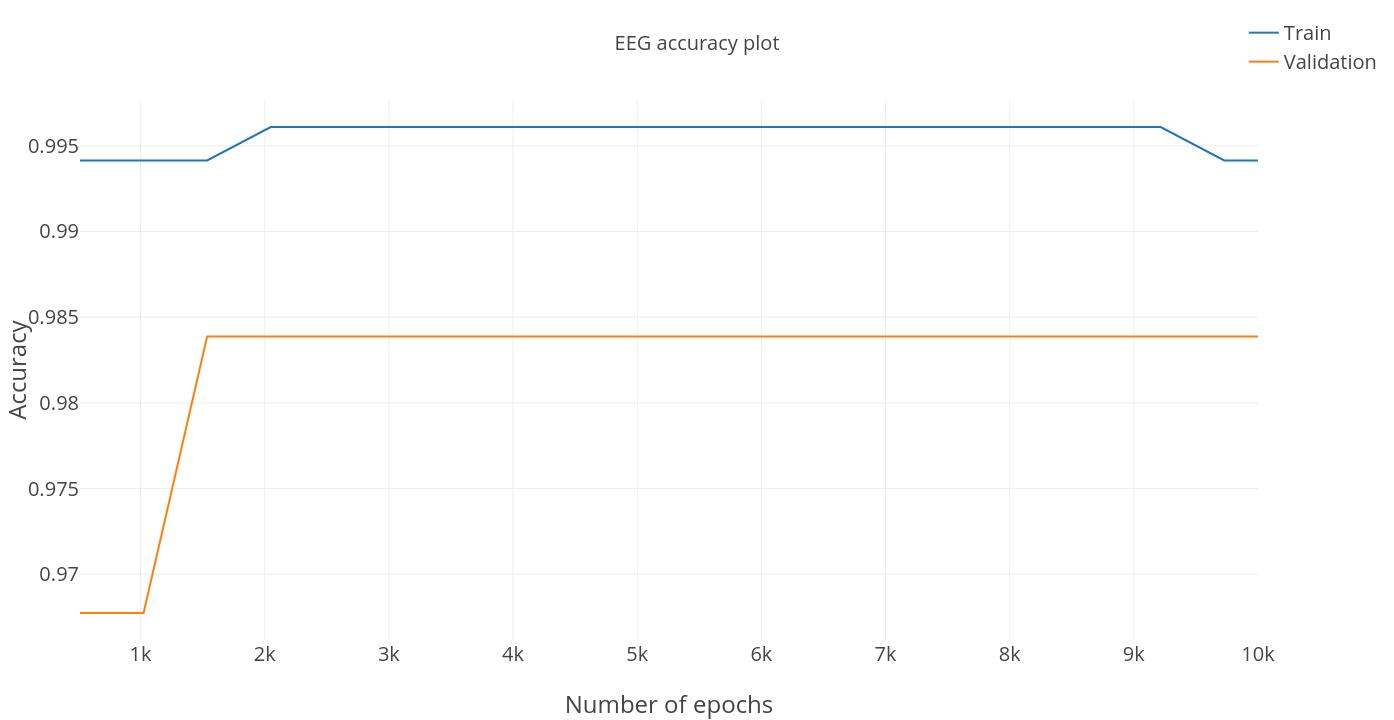}
\caption{ASR using EEG only accuracy plot for recognition of words in presence of background noise} 
\label{1vsall}
\end{figure}
\begin{figure}[!ht]
\includegraphics[width=0.45
\textwidth,trim={0.1cm 2cm 0.1cm 4.5cm},clip]{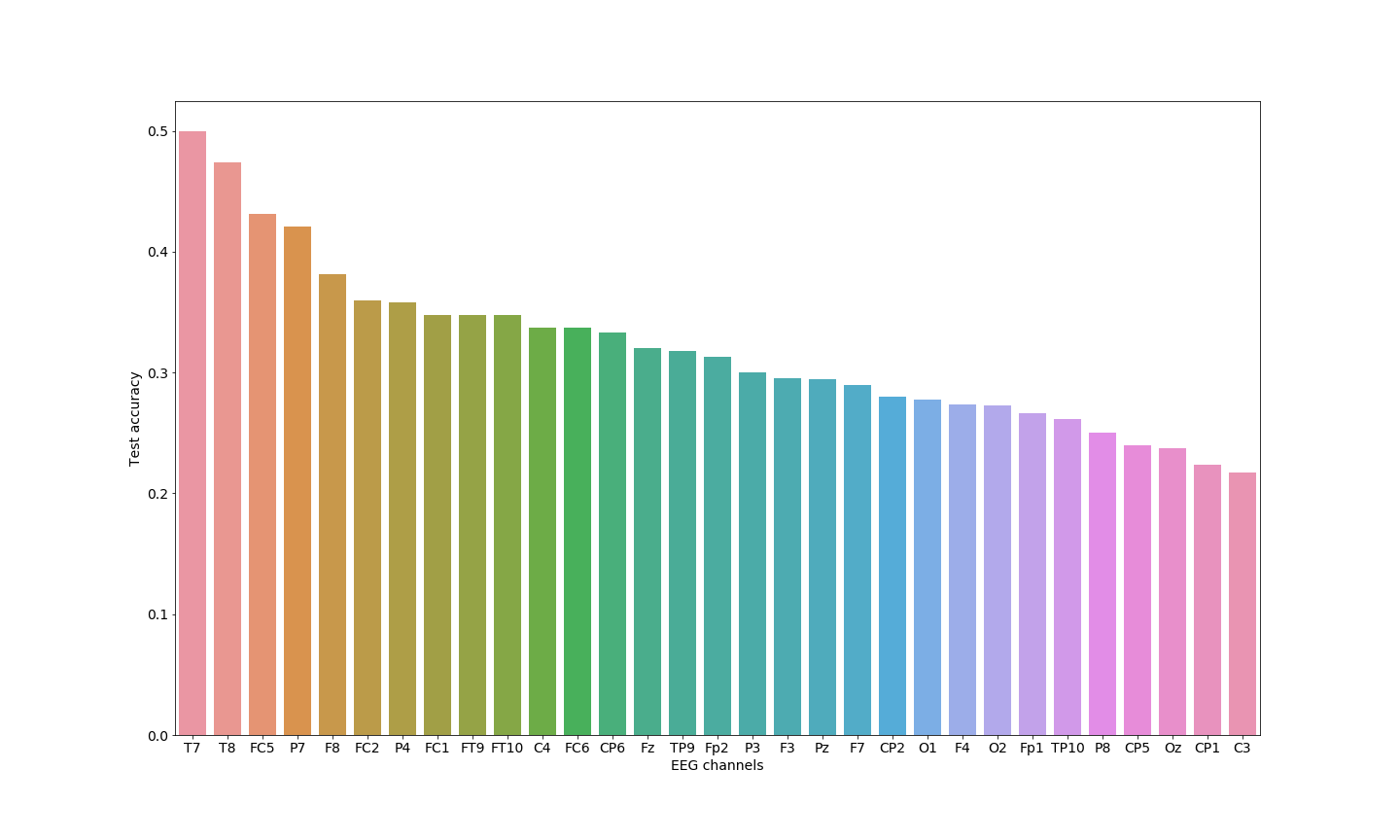}
\caption{ASR test accuracy (EEG only) contribution per each EEG sensor. Channels \textbf{T7, T8, Fc5} and \textbf{P7} showed highest contribution to test accuracy. Dataset used here was vowels in absence of noise.} 
\label{1vsall}
\end{figure}
\begin{table}[!ht]
\begin{center}
    \begin{tabular}{|p{2cm}|p{1.75cm}|p{1cm}|p{1.25cm}|p{1cm}|}
         \hline
 Words/Vowels & Background noise & MFCC acc & MFCC-EEG acc & EEG acc \\ 
          \hline\hline
          Vowels & No & 89.09 & \bf{96.36} & 90.91\\
          \hline
          Vowels & Yes & 74.74 & \bf{94.74} & 92.63\\
          \hline
          Words & No & 95.63 & \bf{97.91} & 96.87\\
          \hline
          Words & Yes & 93.00 & 97.50 & \bf{99.38}\\
          \hline
   
    \end{tabular}
    \caption{Test accuracy for \textbf{ASR EEG fusion} for different datasets}
\end{center}
\end{table}
\begin{table}[!ht]
\begin{center}
    \begin{tabular}{|p{1cm}|p{1cm}|p{1.5cm}|p{1.5cm}|p{1cm}|}
    \hline
    Temp & Lambda & Student Training acc & Student Validation acc & Student Testing acc \\
    \hline \hline
     1  &  0.0 & 99.39 & 97.22 & 97.22 \\
     \hline
    2  & 0.2  & \bf{99.54} & \bf{97.22} & \bf{98.61} \\
    \hline
    5 & 0.8 & 99.23 & 95.83 & 98.61 \\
    \hline
    10 & 1 & 94.94 & 95.83 & 94.44 \\
    \hline
    \end{tabular}
    \caption{Hyper parameter tuning table for \textbf{distillation training}. The data set used here was words with no noise}
\end{center}
\end{table}
\begin{table}[!ht]
\begin{center}
    \begin{tabular}{|p{2cm}|p{1.75cm}|p{1.5cm}|p{1.5cm}|}
         \hline
 Words/Vowels & Background noise & Student acc & MFCC acc  \\ 
          \hline\hline
          Vowels & No & \bf{92.73} & 89.09 \\
          \hline
          Vowels & Yes & \bf{76.84} & 74.74 \\
          \hline
          Words & No & \bf{98.61} & 95.83 \\
          \hline
          Words & Yes & \bf{97.62} & 93.00 \\
          \hline
    \end{tabular}
    \caption{Test accuracy after \textbf{distillation training} for different datasets}
\end{center}
\end{table}

\section{{\bf \uppercase{Conclusion}}}
\label{experiments}
To our knowledge, this is the first time that a deep learning model based speech recognition is demonstrated with high accuracy using only EEG features for character or word level prediction. Our work also demonstrates the ability of EEG to make up for ASR performance lost due to background noise. 
\\
We also show that distillation training can improve the accuracy of an ASR system fused with EEG features. 
We are currently working towards speech recognition for a larger speech EEG corpus. We believe speech recognition using EEG will help people with speaking difficulties to have better technology accessibility.

\bibliographystyle{IEEEbib}

\bibliography{refs} 

\begin{thebibliography}{10}

\bibitem{kirchhoff2002combining}
Katrin Kirchhoff, Gernot~A Fink, and Gerhard Sagerer,
\newblock ``Combining acoustic and articulatory feature information for robust
  speech recognition,''
\newblock {\em Speech Communication}, vol. 37, no. 3-4, pp. 303--319, 2002.

\bibitem{liu2018speech}
Yichuan Liu and Hasan Ayaz,
\newblock ``Speech recognition via fnirs based brain signals,''
\newblock {\em Frontiers in Neuroscience}, vol. 12, pp. 695, 2018.

\bibitem{ramsey2017decoding}
NF~Ramsey, E~Salari, EJ~Aarnoutse, MJ~Vansteensel, MB~Bleichner, and
  ZV~Freudenburg,
\newblock ``Decoding spoken phonemes from sensorimotor cortex with high-density
  ecog grids,''
\newblock {\em Neuroimage}, 2017.

\bibitem{martin2016word}
Stephanie Martin, Peter Brunner, I{\~n}aki Iturrate, Jos{\'e} del~R Mill{\'a}n,
  Gerwin Schalk, Robert~T Knight, and Brian~N Pasley,
\newblock ``Word pair classification during imagined speech using direct brain
  recordings,''
\newblock {\em Scientific reports}, vol. 6, pp. 25803, 2016.

\bibitem{yoshimura2016decoding}
Natsue Yoshimura, Atsushi Nishimoto, Abdelkader~Nasreddine Belkacem, Duk Shin,
  Hiroyuki Kambara, Takashi Hanakawa, and Yasuharu Koike,
\newblock ``Decoding of covert vowel articulation using electroencephalography
  cortical currents,''
\newblock {\em Frontiers in neuroscience}, vol. 10, pp. 175, 2016.

\bibitem{kumar2018envisioned}
Pradeep Kumar, Rajkumar Saini, Partha~Pratim Roy, Pawan~Kumar Sahu, and
  Debi~Prosad Dogra,
\newblock ``Envisioned speech recognition using eeg sensors,''
\newblock {\em Personal and Ubiquitous Computing}, vol. 22, no. 1, pp.
  185--199, 2018.

\bibitem{sun2016neural}
Pengfei Sun and Jun Qin,
\newblock ``Neural networks based eeg-speech models,''
\newblock {\em arXiv preprint arXiv:1612.05369}, 2016.

\bibitem{sriram2018robust}
Anuroop Sriram, Heewoo Jun, Yashesh Gaur, and Sanjeev Satheesh,
\newblock ``Robust speech recognition using generative adversarial networks,''
\newblock in {\em 2018 IEEE International Conference on Acoustics, Speech and
  Signal Processing (ICASSP)}. IEEE, 2018, pp. 5639--5643.

\bibitem{mcloughlin2015robust}
Ian McLoughlin, Haomin Zhang, Zhipeng Xie, Yan Song, and Wei Xiao,
\newblock ``Robust sound event classification using deep neural networks,''
\newblock {\em IEEE/ACM Transactions on Audio, Speech, and Language
  Processing}, vol. 23, no. 3, pp. 540--552, 2015.

\bibitem{gemmeke2011exemplar}
Jort~F Gemmeke, Tuomas Virtanen, and Antti Hurmalainen,
\newblock ``Exemplar-based sparse representations for noise robust automatic
  speech recognition,''
\newblock {\em IEEE Transactions on Audio, Speech, and Language Processing},
  vol. 19, no. 7, pp. 2067--2080, 2011.

\bibitem{tan2018adaptive}
Tian Tan, Yanmin Qian, Hu~Hu, Ying Zhou, Wen Ding, and Kai Yu,
\newblock ``Adaptive very deep convolutional residual network for noise robust
  speech recognition,''
\newblock {\em IEEE/ACM Transactions on Audio, Speech, and Language
  Processing}, vol. 26, no. 8, pp. 1393--1405, 2018.

\bibitem{hinton2014distilling}
Geoffrey Hinton, Oriol Vinyals, and Jeff Dean,
\newblock ``Distilling the knowledge in a neural network,''
\newblock 2014.

\bibitem{yu2016articulatory}
Jianguo Yu, Konstantin Markov, and Tomoko Matsui,
\newblock ``Articulatory and spectrum features integration using generalized
  distillation framework,''
\newblock in {\em Machine Learning for Signal Processing (MLSP), 2016 IEEE 26th
  International Workshop on}. IEEE, 2016, pp. 1--6.

\bibitem{schultz2017biosignal}
Tanja Schultz, Michael Wand, Thomas Hueber, Dean~J Krusienski, Christian Herff,
  and Jonathan~S Brumberg,
\newblock ``Biosignal-based spoken communication: A survey,''
\newblock {\em IEEE/ACM Transactions on Audio, Speech, and Language
  Processing}, vol. 25, no. 12, pp. 2257--2271, 2017.

\bibitem{alsaleh2016brain}
Mashael~M AlSaleh, Mahnaz Arvaneh, Heidi Christensen, and Roger~K Moore,
\newblock ``Brain-computer interface technology for speech recognition: A
  review,''
\newblock in {\em Signal and Information Processing Association Annual Summit
  and Conference (APSIPA), 2016 Asia-Pacific}. IEEE, 2016, pp. 1--5.

\bibitem{rosinova2017voice}
Marianna Rosinov{\'a}, Martin Lojka, J{\'a}n Sta{\v{s}}, and Jozef Juh{\'a}r,
\newblock ``Voice command recognition using eeg signals,''
\newblock in {\em ELMAR, 2017 International Symposium}. IEEE, 2017, pp.
  153--156.

\bibitem{kim2014eeg}
Jongin Kim, Suh-Kyung Lee, and Boreom Lee,
\newblock ``Eeg classification in a single-trial basis for vowel speech
  perception using multivariate empirical mode decomposition,''
\newblock {\em Journal of neural engineering}, vol. 11, no. 3, pp. 036010,
  2014.

\bibitem{chung2014empirical}
Junyoung Chung, Caglar Gulcehre, KyungHyun Cho, and Yoshua Bengio,
\newblock ``Empirical evaluation of gated recurrent neural networks on sequence
  modeling,''
\newblock {\em arXiv preprint arXiv:1412.3555}, 2014.

\bibitem{delorme2004eeglab}
Arnaud Delorme and Scott Makeig,
\newblock ``Eeglab: an open source toolbox for analysis of single-trial eeg
  dynamics including independent component analysis,''
\newblock {\em Journal of neuroscience methods}, vol. 134, no. 1, pp. 9--21,
  2004.

\bibitem{sharbrough1991american}
Frank Sharbrough,
\newblock ``American electroencephalographic society guidelines for standard
  electrode position nomenclature,''
\newblock {\em J clin Neurophysiol}, vol. 8, pp. 200--202, 1991.

\bibitem{zhang2008feature}
Aihua Zhang, Bin Yang, and Ling Huang,
\newblock ``Feature extraction of eeg signals using power spectral entropy,''
\newblock in {\em 2008 International Conference on BioMedical Engineering and
  Informatics}. IEEE, 2008, pp. 435--439.

\end{thebibliography}
\end{document}